\documentclass[latterpaper]{article}
\usepackage{spconf,amsmath,graphicx}

\usepackage{microtype}
\usepackage{subfigure}
\usepackage{booktabs}
\usepackage{hyperref}

\usepackage[utf8]{inputenc}
\usepackage{amssymb, latexsym}
 \usepackage{multirow}

\usepackage{tikz}
\usepackage{xcolor}
\usepackage{authblk}
\usetikzlibrary{decorations.pathreplacing}
\definecolor{fc}{HTML}{1E90FF}
\definecolor{conv}{HTML}{FFA500}
\definecolor{pool}{HTML}{B22222}
\tikzset{fc/.style={black,draw=fc,thick,rectangle,minimum height=0.75cm}}
\tikzset{conv/.style={black,draw=conv,thick,rectangle,minimum height=0.75cm}}
\tikzset{pool/.style={black,draw=pool,thick,rectangle,minimum height=0.75cm}}
     
\title{Sentiment-Aware Automatic Speech Recognition pre-training for enhanced Speech Emotion Recognition}

\name{
Ayoub Ghriss\sthanks{Work was done when Ayoub was working as an intern at Amazon},
Bo Yang$^+$, Viktor Rozgic$^+$, Elizabeth Shriberg\sthanks{Work was done while at Amazon}, Chao Wang$^+$}
\address{$^*$ University of Colorado Boulder, ayoub.ghriss@colorado.edu, $^\dagger$ Ellipsis Health, elizabeth.shriberg@gmail.com, $^+$ Amazon Alexa, \{amzbyang, rozgicv, wngcha\}@amazon.com}

\begin{document}

\maketitle

\begin{abstract}
We propose a novel multi-task pre-training method for Speech Emotion Recognition (SER). We pre-train SER model simultaneously on Automatic Speech Recognition (ASR) and sentiment classification tasks to make the acoustic ASR model more ``emotion aware''. We generate targets for the sentiment classification using text-to-sentiment model trained on publicly available data. Finally, we fine-tune the acoustic ASR on emotion annotated speech data. 
We evaluated the proposed approach on MSP-Podcast dataset, where we achieved the best reported concordance correlation coefficient (CCC) of 0.41 for valence prediction.
\end{abstract}
\begin{keywords}
Speech emotion recognition, automatic speech recognition, sentiment analysis, pre-training
\end{keywords}
\section{Introduction}
%


The origins of Speech Emotion Recognition (SER) date back to Blanton's work \cite{voicemotion} when he wrote that “the language of the tones is the oldest and most universal of all our means of communication.”, and culminated in studies on word-free voice samples with the goal of isolating qualitative features of the voice from those related to the articulated sounds patterns \cite{wordfreemo}. Emotion researchers  define emotion either as a discrete construct or using emotion dimensions. The discrete construction theory credited to Ekman\cite{ekman} isolates six basic emotion categories: anger, disgust, fear, happiness, sadness, surprise, but more comprehensive emotion category sets have been proposed \cite{categories27}. The dimensional perspective, on the other hand, defines emotion as a point in a space defined by emotion dimensions, most commonly Activation, Valence, and Dominance (AVD).

A century later, the challenge of SER remains but new computational tools have emerged permitting more complex modelling. The motivations have also changed, it is no longer destined to understanding the emotion from a psychological perspective alone. It is further fueled by the ubiquitous speech-based interactions. Moreover, building smart systems capable of detecting the user emotional state has the potential of enhancing the interactive experience with different devices.

(Deep) neural network-based models have been a popular choice for SER in recent years \cite{fayek2016correlation, kowtha2020detecting, mao}. However, training a Neural Network for SER requires a large training corpus, and non-neutral emotions are rare in speech. Recent work has addressed this limitation by pre-training on speech tasks, such as ASR \cite{fayek2016correlation, lakomkin2018reusing} or Speaker Identification \cite{bancroft2019exploring, pappagari2020x}.  In this work, we propose a novel pre-training approach for SER. The proposed pre-training consists of building a “Sentiment-aware ASR” (SA2SR). The training of SA2SR objective is a combination of ASR and text-based sentiment classification, where the text sentiment labels are generated from a trained text sentiment model. This approach allows us to amass a large amount of text sentiment labels for speech data, and the effectiveness of these labels on improving SER is validated in our experiments.

\section{Related work}

A suitable choice of input representation is crucial for Speech Emotion Recognition (SER). Traditional approaches used classical signal processing methods (pitch, filter banks) or statistical measures (mean, variance, quantiles...) of the acoustic signal to train on classification/regression (Ekman categories vs AVD). Recent work on SER attempted to infer the emotion based on acoustic and textual cues, either simultaneously or separately.

In End-to-End SER with ASR \cite{e2easr}, an acoustic-to-word ASR model is first trained then fine-tuned on a multi-task learning to jointly optimize ASR and SER objective. The emotion prediction block has two input fields: acoustic features similar to those used in ASR and the states of the ASR decoder. The authors also show that using this combination (raw acoustic inputs \& ASR decoder features) outperforms an SER based on any single element of this combination.

Combining ASR and SER also outperforms a variant in which the SER block model takes the transcript of the ASR (word embeddings) instead of the decoder states. This result is expected since using the ASR transcript propagates the transcription inaccuracies to the SER block. The same reasoning can applied to text-based SER to point out their limitations, such as the one used in Sentiment Analysis based on speaker data \cite{asrsent}. In this speech-based Sentiment Analysis a pre-trained ASR is used for transcription of the utterance. The text is then fed to a feature extractor in parallel to a \textit{Speaker Identification} feature block to provide an input to the emotion predictor.

A different approach that decouples text and acoustic features was introduced in Multi-modal SER \cite{multimodalser}, where the SER model encodes information from audio and text using two encoders. For the text encoder, the text input is the transcription from an ASR. This approach leverages the ability of text encoders to capture long term semantics better than the acoustic ones. However, this multi-modal approach assumes that the transcript is provided (via ASR), which limits its applicability when only the utterance (audio) is accessible.


To the extent of our knowledge, the only previous work that leveraged text-based sentiment labels was published recently\cite{LanguageModelASAAR} with three major differences: 1) we start with analyzing the correlation between text sentiment and speech emotion, thereby establishing a strong motivation for the proposed method, and an explanation for the observed performance boost, 2) the focus of this work is on SER with the widely used dimensional emotion representation (activation, valence and dominance), while that of \cite{LanguageModelASAAR} is on three-way speech sentiment classification, and 3) The approach in \cite{LanguageModelASAAR} uses out of the box encodings (ASR followed by Bert) and is oblivious to the feedback from sentiment labels. Our approach, on the other hand, leverages the proxy sentiment labels to induce the ASR embedding to incorporate emotional knowledge and yields better performance.
\section{Proposed method}
We propose building a SER model that is pre-trained on a sentiment-aware ASR task. The sentiment-awareness is implemented by transferring sentiment knowledge from a text domain to the acoustic domain.

\subsection{Correlation between text sentiment and speech emotion}
\label{sec:example}
We conjecture that text sentiment correlates with the valence dimension of speech emotion. Indeed, when a human listener tries to determine the emotion from a speech segment, the words in the speech also plays a role -- the obvious examples are the speech segments that contain cursing words and phrases, or strong praising adjectives (e.g. excellent, beautiful, etc).

We test this correspondence between text-based sentiment and valence on the IEMOCAP dataset \cite{busso2008iemocap} and use a pre-trained text sentiment analysis (Roberta \cite{sentimentroberta}) to get the sentiment labels: \textit{ negative,  positive,  neutral}. Table \ref{fig:confusion} shows the confusion matrix between text sentiment classes and speech emotion, with the dominant speech emotion highlighted in red in each sentiment class and the second dominant ones highlighted in blue.

As can be seen from Table \ref{fig:confusion}, the negative text-sentiment utterances are mostly associated with negative speech emotion labels (Sad, Anger and Frustrated); while the positive text-sentiment utterances correspond to positive speech emotion labels (Happy). More interestingly, by investigating elements of the cell $\{Neutral,  frustrated\}$ we can find transcripts such as : \textit{``Nothing'',  ``A vacation.'',  ``I'm just saying.''}$\}$, while the inferred sentiment is neutral -- this means the emotion in this case is likely conveyed through speech style and tone. Furthermore, by grouping 
$\{Sad, Frustrated, Anger\}$ into one class, we get a Spearman correlation of $0.22$ between text sentiment and speech emotion. These observations motivate us to employ readily available text sentiment model to generate sentiment labels, which will serve as weak signal to train speech emotion models.

\begin{table}[h]
\centering
\resizebox{0.5\textwidth}{!}{
\begin{tabular}{llccc}
\toprule
\multicolumn{2}{c}{}  & \multicolumn{3}{c}{Text sentiment} \\
\multicolumn{2}{c}{} & Negative & Neutral & Positive \\
 \midrule
\multirow{5}{*}{Speech emotion} & Sad &  339 & 604 & 137 \\
& Anger & \textbf{\color{blue}490} & 518 &  94 \\
& Frustrated & \textbf{\color{red}658} & \textbf{\color{blue} 1049} & 141 \\
& Neutral & 253 & \textbf{\color{red} 1251} & \textbf{\color{blue}204} \\
& Happy & 252 & 848 & \textbf{\color{red} 533} \\
\bottomrule
\end{tabular}
}
\caption{Confusion matrix between text-based sentiment \cite{sentimentroberta} and speech emotion (IEMOCAP)\label{fig:confusion}}
\end{table}
\subsection{Sentiment-aware Automatic Speech Recognition (SA2SR)}

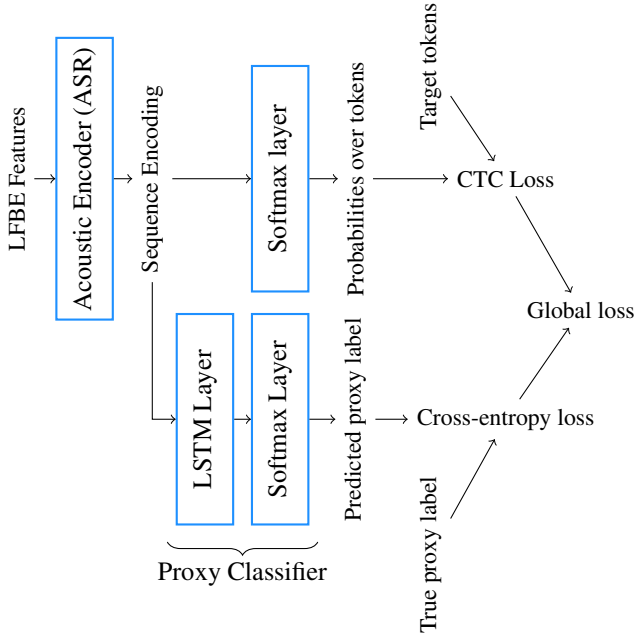
\begin{figure}[h]
\centering
\begin{tikzpicture}
    \node[rotate=90] (lfbe) at (-2,0) {\small LFBE Features};
    \node[fc,rotate=90,minimum width=3cm] (lstm) at (-1.1,0) {Acoustic Encoder (ASR)};
    \node[rotate=90] (encode) at (-0.2,0) {\small Sequence Encoding};
    \node[fc,rotate=90,minimum width=3cm] (soft) at (1.5,0) {Softmax layer};
    \node[rotate=90] (tokens) at (2.5,0) {\small Probabilities over tokens};
    \node[rotate=90] (truetokens) at (3.5,1.5) {\small Target tokens};
    \node[rotate=0] (ctc) at (4.5,0) {\small CTC Loss};

    \node[fc,rotate=90,minimum width=2.8cm] (lstm2) at (0.5,-3.2) {LSTM Layer};
    \node[fc,rotate=90,minimum width=2.8cm] (soft2) at (1.5,-3.2) {Softmax Layer};
    \node[rotate=90] (sent) at (2.5,-3.2) {\small Predicted proxy label};
    \draw [decorate,decoration={brace,amplitude=5pt,mirror,raise=4ex}]
  (0,-4.2) -- (2,-4.2) node[midway,yshift=-3em]{Proxy Classifier};
    \node[rotate=90] (truesent) at (3.5,-5) {\small True proxy label};
    \node[rotate=0] (cross) at (4.5,-3.2) {\small Cross-entropy loss};
    \node[rotate=0] (global) at (5.5,-1.75) {\small Global loss};
    
    \draw[->] (lfbe) edge (lstm);
    \draw[->] (lstm) edge (encode);
    \draw[->] (encode) edge (soft);
    \draw[->] (soft) edge (tokens);
    \draw[->]  (tokens) edge (ctc);
    \draw[->]  (truetokens) edge (ctc);
    
    \draw[->,to path={|- (\tikztotarget)}]  (encode) edge (lstm2);
    \draw[->] (lstm2) edge (soft2);
    \draw[->] (soft2) edge (sent);
    \draw[->]  (truesent) edge (cross);
    \draw[->]  (sent) edge (cross);
    \draw[->]  (cross) edge (global);
    \draw[->]  (ctc) edge (global);
\end{tikzpicture}
\caption{Architecture of SA2SR pre-training network}
\label{fig:network}
\end{figure}

In addition to the ASR model, a sentiment classifier (proxy classifier) is trained jointly on the acoustic encoder states. The architecture logic is similar to the one in combined ASR-SER\cite{e2easr}. The model takes as input the log filterbank energy (LFBE) features and contains two classifiers (Figure ~\ref{fig:network}) that take the encoded acoustic sequence as input:
\begin{itemize}
    \item Sequence to sequence classifier: A softmax layer for token classification
    \item Sentiment classifier: A sequence summarizer (recurrent neural network) followed by a softmax layer over the sentiment classes (\textit{negative, neutral, positive}).
\end{itemize}
The proposed architecture is trained using a loss that is a linear combination (Equation ~\ref{eq:loss}) of (a) Connectionist Temporal Classification (CTC) loss \cite{graves2006connectionist} between the target sequence and the sequence of output probabilities over the token set, and (b) cross entropy loss between the predicted and proxy sentiment targets, i.e., sentiment obtained from the pre-trained text-to-sentiment model. The global loss is defined as:
\begin{align}
    L_{global} = L_{ASR} + \lambda L_{sentiment}
    \label{eq:loss}
\end{align}
where $\lambda \geq 0$ is a hyper-parameter reflecting how the importance of sentiment classification vis-\`a-vis the ASR task.

\subsection{Fine-tuning}

\begin{figure}
\centering
\begin{tikzpicture}
    \node[rotate=90] (lfbe) at (-1,0) {\small LFBE Features};
    \node[pool,rotate=90,minimum width=3cm] (lstm) at (-0.1,0) {Acoustic Encoder (ASR)};
    \node[rotate=90] (encode) at (0.8,0) {\small Sequence Encoding};
    \draw [decorate,decoration={brace,amplitude=5pt,mirror,raise=4ex}]
  (-0.1,-1.5) -- (1,-1.5) node[midway,yshift=-3em]{Pre-trained};
    \node[fc,rotate=90,minimum width=3.5cm] (conv1d) at (1.8,0) {1-D Convolution Block};
    \node[fc,rotate=90,minimum width=3.5cm] (attn) at (2.8,0) {Multi-head attention};
    \node[fc,rotate=90,minimum width=3.5cm] (pool) at (3.8,0) {$\sigma$ and $\mu$ pooling};
    \node[fc,rotate=90,minimum width=3.5cm] (soft2) at (4.8,0) {Linear Projection};
    \node[rotate=90] (sent) at (5.6,0) {\small Predicted AVD};
    \node[rotate=90] (truesent) at (6.8,-1.5) {\small True AVD};
    \node[rotate=0] (ccc) at (6.8,0) {\small CCC loss};
    \draw [decorate,decoration={brace,amplitude=5pt,mirror,raise=4ex}]
  (1.8,-1.5) -- (4.8,-1.5) node[midway,yshift=-3em]{Emotion Regressor};
    
    \draw[->] (lfbe) edge (lstm);
    \draw[->] (lstm) edge (encode);
    \draw[->,]  (encode) edge (conv1d);
    \draw[->,]  (conv1d) edge (attn);
    \draw[->] (attn) edge (pool);
    \draw[->] (pool) edge (soft2);
    \draw[->] (soft2) edge (sent);
    \draw[->]  (truesent) edge (ccc);
    \draw[->]  (sent) edge (ccc);
\end{tikzpicture}
\caption{Architecture of the fine-tuned SA2SR}
\label{fig:finetuning}
\end{figure}
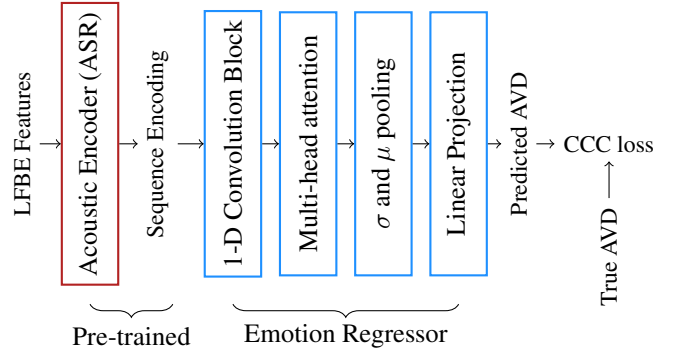

During fine-tuning, we use the pre-trained acoustic encoder (Figure~\ref{fig:finetuning}) and add an emotion regression transformer block that takes the encoding sequence as input and outputs the Activation, Valence, Dominance (AVD) values. The model is then trained to maximize the Concordance Correlation Coefficient (CCC) between prediction $\hat{y}$ and target values $y$.
\begin{align}
CCC(y,\hat{y}) = \frac{2 Cov(y,\hat{y})}{\sigma^2_y + \sigma^2_{\hat{y}} + (\mu_y - \mu_{\hat{y}})^2}    
\end{align}
where $Cov$ denotes the co-variance, $\sigma$ and $\mu$ denote the sample variance and mean, respectively. During model training, these statistics are computed on mini-batches of data.

To enable easy comparison to previous works, the CCC objectives for each of the activation, valence and dominance are simply combined as:
\begin{align}
    \mathcal{L_{CCC}} = - \frac{1}{3}(CCC_{A} + CCC_{V} + CCC_{D})
\end{align}

\section{Experiment results}
Datasets: We use the full Librispeech dataset \cite{librispeech} for pretraining a character based SA2SR. The dataset contains audios and transcripts for around 960 hours of audio, and we generate the proxy sentiment labels using text-to-sentiment RoBerta model trained on Twitter sentiment data \cite{Heitmann2020MoreTA}. We extract LFBE features with 40 frequencies using $25ms$ window and $10ms$ steps. Hence, for SA2SR pre-training, an input tuple consists of \textit{(LFBE Features, transcript, proxy label)}.\par

Model fine-tuning and evaluation are conducted on MSP-Podcast \cite{MSP} dataset. The MSP-Podcast contains ~60K speech segments from podcast recordings which are perceptually annotated using crowd-sourcing. The ground truth emotion dimension scores are obtained averaging scores selected by individual annotators on seven-point \textit{Likert} scale. The dataset was split following the same partitions provided in the official MSP corpus.

\subsection{Networks Architecture}
The Acoustic Encoder consists of 5 Bidirectional LSTM (Bi-LSTM) layers. All LSTM layers have 192 units with $tanh$ activation and $sigmoid$ recurrent activation. 

The 1-D convolutional block in Emotion Regressor (Figure ~\ref{fig:finetuning}) is a stack of 2 one-dimensional \textit{masked convolutions} with filter sizes (6,3) and strides (3,2) respectively. The convolutions accept input masks and process them accordingly to match the output of the convolution. The convolutions are followed by sample level normalization and a $\text{LeakyRELU}_{\alpha=0,3}$ is applied to the convolutional block output. 


The multi-head attention uses 4 attention heads and 64 dimensions for all encoding spaces and feed-forward outputs. The output of the attention mechanism is pooled by computing the variance and the mean of the sequences yielding 128 features that are linearly projected into the AVD space.

    
    
    
    

\subsection{Pre-training}
For the global loss weight, we chose $\lambda = 200$. To optimize it, we use Adam optimizer with the parameters: $lr=5 \times 10^{-5}$, $\beta_1=0.9, \beta_2=0.999$. The $\lambda$ was chosen such that the ASR and sentiment classification losses achieve similar value on the validation set.

The LFBE features are normalized on a sample level. The training set is then augmented using speed augmentation of factors $0.9$ and $1.1$. This augmentation step triples the training set size. The features are then masked on time and frequency dimensions with probability $p=0.5$, as in the SpecAugment method \cite{specaug}. 
We finally stack every adjacent three frames and skip every other two time steps, which leads to shorter sequences and improved model training time. The ASR training uses a token set of 29 characters.

After each epoch, we use the validation set to compute the Character Error Rate (CER) from the ASR task and Area Under the Curve (AUC) of the sentiment classifier. The pre-training terminates when the metric $M = CER - AUC$ does not improve for 25 epochs and the model that has the lowest $M$ is evaluated. For the baseline model, we pre-train on the ASR task only and use the model from the epoch with the best CER on the validation set.

\subsubsection{Effectiveness of ASR features for SER}
To start, we test the effectiveness of ASR trained features for SER. Some previous work, for example \cite{fayek2016correlation, lakomkin2018reusing}, reported limited transferability between ASR and SER. In our experiments, however, we found ASR features to be quite effective. It is possible that the differences are due to the large pre-training dataset (full Librispeech) and modern model architectures such as bidirectional LSTM and transformers.

\begin{table}[ht]
\centering
\begin{tabular}{lccc}
\toprule
     Model & Activation & Valence & Dominance \\
     \midrule
     No Pre-training &	0.603&	0.304&	0.511\\
ASR features &	0.649& \textbf{	0.393}&	0.544\\
\bottomrule
\end{tabular}
    \caption{The baseline ASR pre-training achieves much improved CCC compared to no pre-training}
    \label{tab:asr_feats}
\end{table}

We report these results in Table \ref{tab:asr_feats}. 
The ``ASR features'' are pre-trained on Librispeech and fine-tuned on MSP-Podcast for SER. The ``No pre-training'' baseline is directly trained with the MSP-Podcast. We observe large performance boost when ASR pre-training is employed.


\subsubsection{Effectiveness of SA2SR pre-training}
In this experiment, we examine whether the additional proxy sentiment labels enhance SER performance. In particular, we expect that valence recognition to be improved, as we have seen correlation between text sentiment and valence in Section \ref{sec:example}. The results are reported in Table \ref{tab:SA2SR_res}. As a comparison, we also included the results from \cite{mao} that uses self-supervised pre-training as a strong contender.

\begin{table}[ht] 
\centering
\resizebox{0.5\textwidth}{!}{
\begin{tabular}{lccc}
\toprule
     Model & Activation & Valence & Dominance \\
     \midrule
ASR features &	0.649& 0.393 &	0.544\\
CPC-based pre-traing \cite{mao} &	\textbf{0.706}&	0.377&	\textbf{0.639}\\
SA2SR features&	0.679&	\textbf{0.412} &	0.564\\
\bottomrule
\end{tabular}
}
\caption{SA2SR produces good features for enhancing the CCC metric}
\label{tab:SA2SR_res}
\end{table}

We can see from Table \ref{tab:SA2SR_res} that recognition of valence, arguably the most important dimension of emotion, is further improved compared to the strong ``ASR features'' baseline. However, in this equal-weight multi-task emotion training setting, we see that activation and dominance dimension performs relatively weak compare to that of \cite{mao}. We view this as an encouraging result as in many applications, valence (positive v.s. negative) is of most interest.
 
Additionally, during the pre-training of the SA2SR model, we observed that the sentiment classification part achieves weighted-average-recall of 0.71 and 0.81 AUC. This indicates that indeed the model is trained to recognize these proxy sentiment labels. Therefore, we expect the learned representation to be suitable for the final SER task.

\subsubsection{Importance of fine-tuning}
Lastly, we are interested in examining whether we should freeze the learned representations during SER training. For both ASR and SA2SR features, we train models with and without freezing the acoustic encoder. The results are reported in Table \ref{tab: pt_vs_ft}.
\begin{table}[h!] 
\centering
\resizebox{0.5\textwidth}{!}{
\begin{tabular}{lccc}
\toprule
Pretraining & Activation & Valence & Dominance \\
\midrule
ASR features (frozen) &	0.503&	0.378&	0.438\\
ASR features &	0.649& 0.393 &	0.544\\
SA2SR features (frozen) &	0.508&	0.403&	0.483\\
SA2SR features &	0.679&	0.412 &	0.564\\
\bottomrule
\end{tabular}
}
\caption{Impact of finetuning the pre-trained encoders on CCC}
\label{tab: pt_vs_ft}
\end{table}

As can be seen from Table \ref{tab: pt_vs_ft}, fine-tuning the encoder gives better SER performance compared to the methods with frozen encoders. This result proves that using out of the box ASR models for transcription without any fine-tuning would be outperformed when the gradient propagates to the ASR network weights. It also proves the point we made earlier about the importance of sentiment awareness in the SA2SR. Even without fine-tuning, SA2SR clearly outperforms the ASR embedding on 2 out of the 3 emotion AVD dimensions.

\section{Conclusion}
We proposed a novel pre-training method that utilizes proxy sentiment labels to aid ASR pre-training for SER. As text-sentiment and speech-emotion are correlated, this way we train speech representations capturing both phonetic and emotion-relevant info. We evaluated the proposed method on the MSP-Podcast dataset achieving state of the art performance on the challenging valence dimension.

Albeit we focused on ASR-based pre-training, the proxy sentiment classification task  can be combined with other pre-training techniques, such as APC \cite{chung2020apc}, CPC \cite{oord2018representation}, which we will address in the future work. 
\newpage
\bibliographystyle{IEEEbib}
\bibliography{ms}

\end{document}